\documentclass[10pt,twocolumn,letterpaper]{article}

\usepackage{cvpr}
\usepackage{color}
\usepackage{times}
\usepackage{epsfig}
\usepackage{graphicx}
\usepackage{amsmath}
\usepackage{amssymb}
\usepackage{multirow}
\usepackage[bold]{hhtensor}
\usepackage{authblk}
\usepackage[pagebackref=true,breaklinks=true,letterpaper=true,colorlinks,bookmarks=false]{hyperref}

\def\cvprPaperID{343} % *** Enter the CVPR Paper ID here

\cvprfinalcopy

\newcommand{\squishlist}{
 \begin{list}{$\bullet$}
  { \setlength{\itemsep}{0pt}
     \setlength{\parsep}{1pt}
     \setlength{\topsep}{1pt}
     \setlength{\partopsep}{0pt}
     \setlength{\leftmargin}{1.5em}
     \setlength{\labelwidth}{1em}
     \setlength{\labelsep}{0.5em} } }

\newcommand{\squishend}{
  \end{list}  }

\begin{document}

\title{Exploit Bounding Box Annotations for Multi-label Object Recognition}

\author[1]{Hao Yang}
\author[2]{Joey Tianyi Zhou}
\author[3]{Yu Zhang}
\author[4]{Bin-Bin Gao}
\author[4]{Jianxin Wu}
\author[1]{Jianfei Cai}
\affil[1]{SCE\\Nanyang Technological University\\ \url{lancelot365@gmail.com}\\\url{ASJFCai@ntu.edu.sg}}
\affil[2]{IHPC\\A*STAR\\ \url{zhouty@ihpc.a-star.edu.sg}}
\affil[3]{Bioinformatics Institute\\A*STAR\\ \url{zhangyu@bii.a-star.edu.sg}}
\affil[4]{National Key Laboratory for Novel Software Technology\\Nanjing University\\China \url{gaobb@lamda.nju.edu.cn}\\\url{wujx2001@nju.edu.cn}}
\maketitle

\begin{abstract}
 Convolutional neural networks (CNNs) have shown great performance as general feature representations for object recognition applications. However, for multi-label images that contain multiple objects from different categories, scales and locations, global CNN features are not optimal. In this paper, we incorporate local information to enhance the feature discriminative power. In particular, we first extract object proposals from each image. With each image treated as a bag and object proposals extracted from it treated as instances, we transform the multi-label recognition problem into a multi-class multi-instance learning problem. Then, in addition to extracting the typical CNN feature representation from each proposal, we propose to make use of ground-truth bounding box annotations (strong labels) to add another level of local information by using nearest-neighbor relationships of local regions to form a multi-view pipeline. The proposed multi-view multi-instance framework utilizes both weak and strong labels effectively, and more importantly it has the generalization ability to even boost the performance of unseen categories by partial strong labels from other categories. Our framework is extensively compared with state-of-the-art hand-crafted feature based methods and CNN based methods on two multi-label benchmark datasets. The experimental results validate the discriminative power and the generalization ability of the proposed framework. With strong labels, our framework is able to achieve state-of-the-art results in both datasets.
\end{abstract}
\thispagestyle{empty}
\section{Introduction} \label{intro}
\begin{figure}
 \centering
 \includegraphics[width=0.38\textwidth]{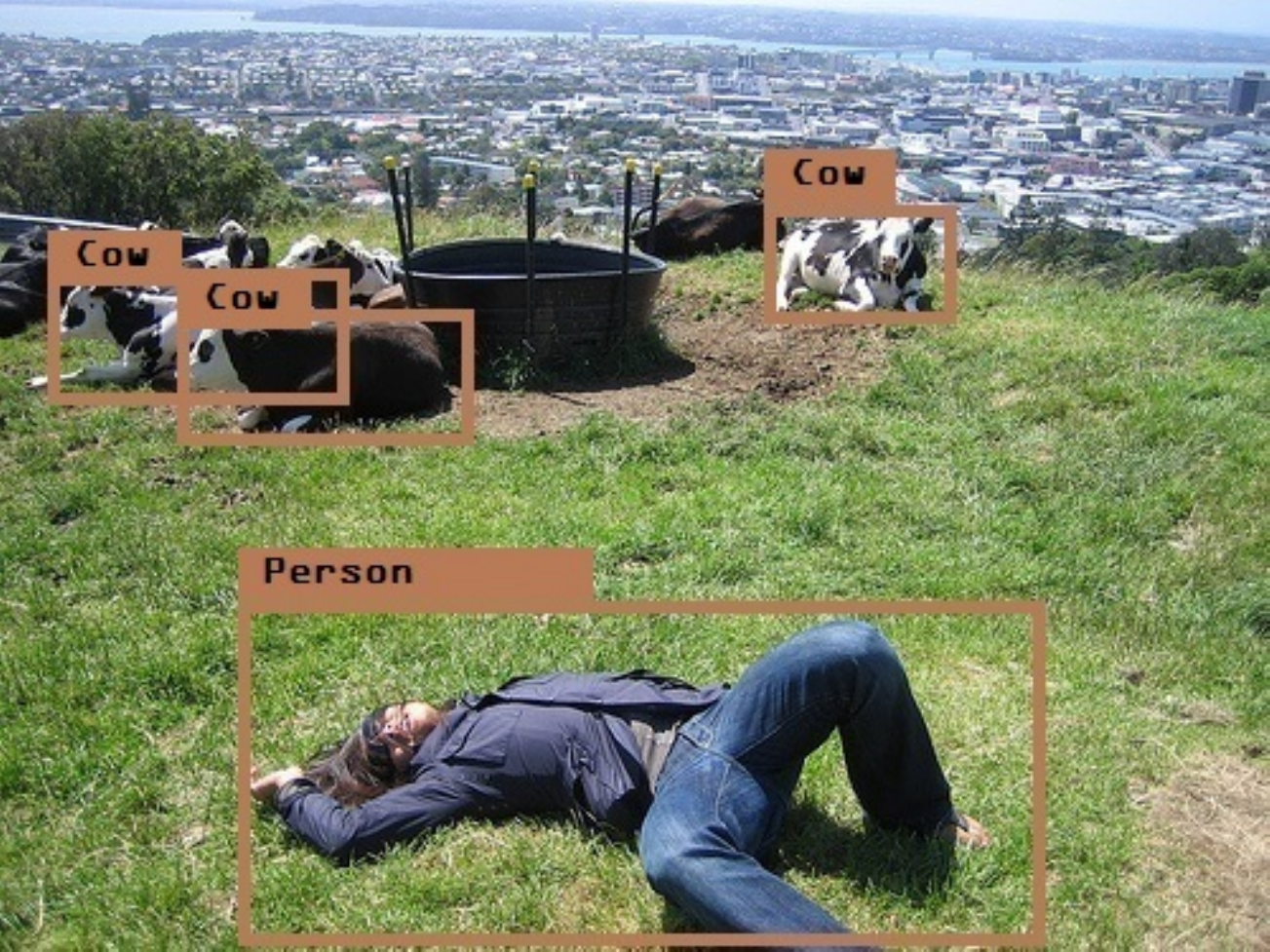}
 \caption{An example of a typical multi-label image, which contains several cows in different locations as well as a person.}
 \label{example-voc}
\end{figure}

Recently, the availability of large amount of labeled data has
greatly boosted the development of feature learning methods for
classification. In particular, convolutional neural networks (CNNs)
achieve great success in visual recognition/classification tasks.
Features extracted from CNNs can provide powerful global
representations for the single object recognition
problem~\cite{Alex2012,Sermanet2013,Razavian2014}. However,
conventional CNN features may not generalize well for images
containing multiple objects as the objects can be in different
locations, scales, occlusions and categories. Fig.~\ref{example-voc} shows an
example of such images. Since multi-label recognition task is more
general and practical in real world applications, many CNN related
methods~\cite{OquabCVPR2014,Oquab2014,Wei2014} have been proposed
to address the problem.

A well known fact is that image-level labels can be utilized to
fine-tune a pre-trained CNN model and produce good global
representations~\cite{Sermanet2013,Razavian2014,Oquab2014}.
However, due to the diversity and the complexity of multi-label
images, classifiers trained from such global representations might
not be optimal. For example, if we use images similar to
Fig.~\ref{example-voc} to train a classifier for ``person'', the
classifier will have to account for not only hundreds of
different variations of ``person'' but also other objects contained in the images. The complexity of
multi-label images adds an extra level of difficulty for training
appropriate classifiers with the global image representations.
Furthermore, due to the large intra-class variations of
multi-label images, the global features extracted from training
images are likely to be unevenly distributed in the feature space.
A classifier trained with such features can be successful at
regions that are densely populated with
training instances, but may fail in poorly sampled areas of the
feature space~\cite{Yu2014}.

\begin{figure*}
  \centering
 \includegraphics[width=0.7\textwidth]{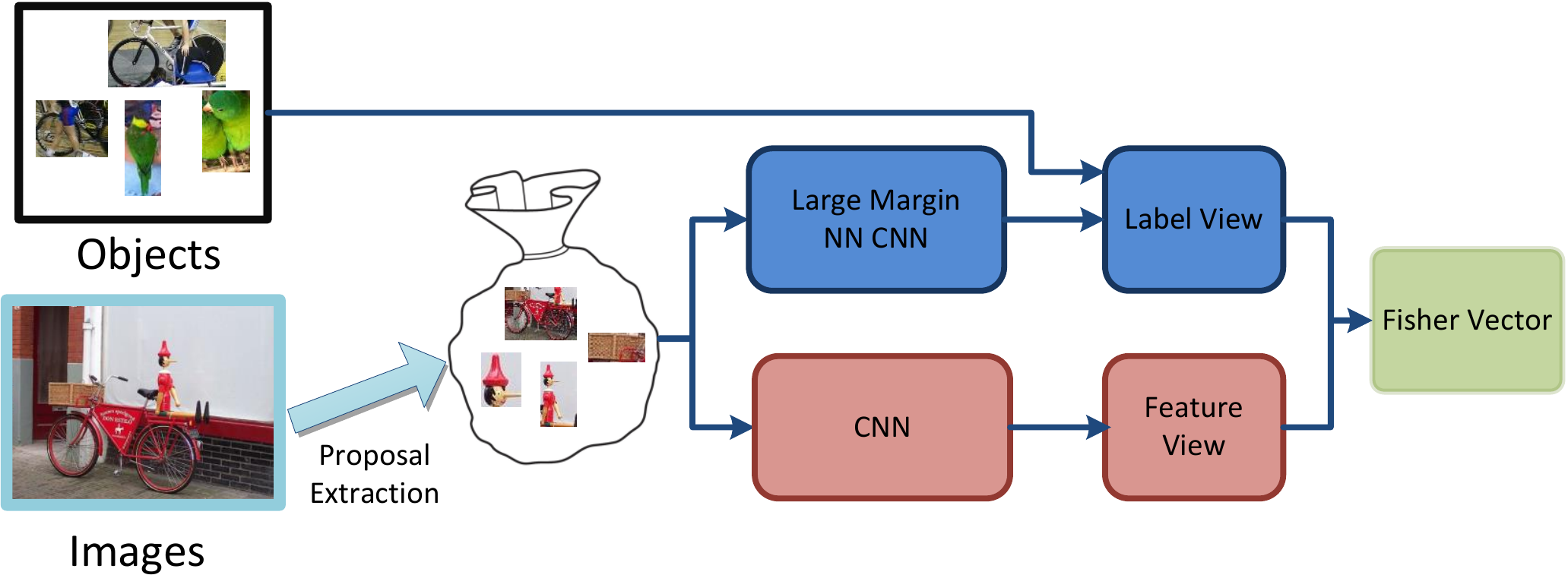}
 \caption{Overview of the proposed multi-view multi-instance framework. We transform the multi-label object recognition problem into a multi-class multi-instance learning problem by first extracting object proposals from each image using selective search. Two types of features are then extracted for each proposal. One is a low-dimensional feature from a large-margin nearest neighbor (LMNN) CNN, which is used to generate the label view by encoding the label information of $k$-NN from the candidate pool (containing ground truth objects). The other is a standard CNN feature as the feature view. These two views are fused and then used to encode a Fisher vector for each image.}
 \label{system}
\end{figure*}

To address the problems with global CNN representations, following the recent works~\cite{Oquab2014,Wei2014,OquabCVPR2014,girshick2014}, we incorporate local information via extracting object proposals using general object detection techniques such as selective search~\cite{Uijlings2013} for multi-label object recognition. By decomposing an image into
local regions that could potentially contain objects, we avoid the complex process
of directly recognizing multiple objects
in the whole image. Instead, we only need to identify whether
there exist target objects in the local regions. However, as the
local regions are noisy and of large variations (see
Fig.~\ref{selective-search}), the usual CNN representations might
not be good enough for discrimination. Therefore, we add another
level of locality by incorporating local nearest-neighbor
relationships of these local regions. In this way, the resulting features will be more evenly
distributed in the feature space. However, such relationships are not easy
to obtain only through weak supervision, i.e., image-level labels.
Fortunately, for many multi-label applications, we can exploit the
strong supervision information, i.e. ground-truth bounding boxes, which can be considered as local regions with strong labels. Then, we can
exploit the relationships between object proposals and ground-truth bounding boxes (e.g.,
nearest neighbor relationships) to help multi-label recognition.

We would like to point out that ground-truth bounding boxes have been utilized in two proposal-based methods~\cite{OquabCVPR2014,girshick2014} for multi-label object recognition. In particular,~\cite{girshick2014} makes use of ground-truth bounding boxes to train category-specific classifiers to classify object proposals. However, it requires ground-truth bounding boxes for each category of objects, which might not be available in practice. In contrast, for our proposed method, even with only partial
strong labels (e.g., bounding boxes and labels for $10$ classes in
the $20$ classes of Pascal VOC), the proposed local relationships
generalize well and can help recognize all classes (e.g., improving
recognition of the $20$ classes in VOC). \cite{OquabCVPR2014} directly uses ground-truth bounding boxes to fine tune the CNN model, but its performance is not better than other proposal-based methods~\cite{Oquab2014,Wei2014} that do not utilize ground-truth bounding boxes. In other words, an effective way to exploit ground-truth bounding boxes for multi-label object recognition is still missing, which is what we aim to provide in this paper.

Fig.~\ref{system} gives an overview of our proposed framework. We
utilize both strong and weak labels as two \emph{views}, and
propose a multi-view multi-instance framework to tackle the
multi-label object recognition task. In particular, for any image, we first
extract object proposals using general object detection
techniques. The global image and its accompanying weak (image)
label is used to fine-tune a standard CNN to generate a \emph{feature view}
representation for each proposal. Using the ground truth bounding
boxes and their strong labels, we design a large margin nearest neighbor
(LMNN) CNN architecture to learn a low-dimensional feature so that
we could extract nearest neighbor relationship between local
regions and a candidate pool formed by ground truth objects. These
local NN features are used as the \emph{label view}. When
combining both views, we can achieve a balance between global
semantic abstraction and local similarity, hence enhancing the
discriminative power of our framework. More importantly, as the
strong labels are indirectly utilized through LMNN to encode local
neighborhood relationships among labelled local regions, the
proposed framework can generalize well to the whole local region
space, even with only partial strong labels for part of the object
classes, making our framework more practical.

The main contribution of this research lies in the proposed multi-view multi-instance framework, which utilizes bounding box annotations (strong labels) to encode the label view and combine it with the typical CNN feature representation (feature view) for multi-label object recognition. Another novelty of our work is the proposed LMNN CNN which effectively extracts local information from the strong labels.

\section{Related Works} \label{related}

Our paper mainly relates to the topics of CNN based multi-label
object recognition, multi-view and multi-instance learning and
local and metric learning.

{\bf CNN based multi-label object recognition}. Recently, CNN
models have been adopted to solve the multi-label object
recognition problem. Many
works~\cite{girshick2014,Sermanet2013,OquabCVPR2014,Oquab2014,Wei2014}
have demonstrated that CNN models pre-trained on a large dataset
such as ILSVRC can be used to extract features for other
applications without enough training data. A typical way
(\cite{Sermanet2013},~\cite{Razavian2014}
and~\cite{Chatfield2014}) is to directly apply a pre-trained CNN
model to extract an off-the-shelf global feature for each image
from a multi-label dataset, and use these features for
classification. However, different from single-object images from
the ImageNet, multi-label images usually have multiple objects in
different locations, scales and occlusions, and thus global
representations are not optimal for solving the
problem~\cite{Wei2014}. More recently, two proposal-based
methods~\cite{OquabCVPR2014,girshick2014} were propose for
multi-label recognition and detection tasks with the help of
ground truth bounding boxes. These methods achieve significant
improvement over single global representations. On the other hand,
\cite{Oquab2014} and \cite{Wei2014} handle the problem in a weakly
supervised manner by max-pooling image scores from the proposal
scores. Moreover,~\cite{Simonyan2014} employs a very deep CNN,
aggregates multiple features from different scales of the image
and achieves state-of-the-art results.

{\bf Multi-view and multi-instance learning}. Multi-view learning
deals with data from multiple sources or feature sets. The goal of
multi-view learning is to exploit the relationship between views
to improve the performance or reduce model complexity. Multi-view
learning is well studied in conjunction with semi-supervised
learning or active learning. To combine information from
multi-views for supervised learning, fusion techniques at feature
level or classifier level can be employed~\cite{Zhang2014}.
Multi-instance learning aims at separating bags containing
multiple instances. Over the years, many multi-instance learning
algorithms have been proposed, including miBoosting~\cite{Xu2004},
miSVM~\cite{Andrews2003}, MILES~\cite{Chen2006} and
miGraph~\cite{Zhou2009}. Several works also studied the
combination of multi-view and multi-instance learning and its
application to computer vision tasks.

{\bf Local and metric learning}. Existing local learning methods
mainly vary in the way that they utilize the labelled instances
nearest to a test instance. One way is to only use a fixed number
of nearest neighbors to the test point to train a model using
neural network, SVM or just voting. The other way is to learn a
transformation of the feature space (e.g., Linear Discriminant
Analysis). In either way, the learned model can be better tailored
for the test instance's neighborhood property~\cite{Hastie1996}.
Metric learning is closely related to local learning as a good
distance metric is crucial for the success of local learning.
Generally, metric learning methods optimize a distance metric to
best satisfy known similarity constraints between training
data~\cite{Bellet2013}. Some metric learning methods learn a
single global metric~\cite{Weinberger2009}. Others learn local
metrics that vary in different regions of the feature
space~\cite{Yang2006}.

\section{Multi-Label as Multi-Instance} \label{mi}
In this section, we introduce the first level of locality by
formulating the multi-label object recognition problem as a
multi-instance learning (MIL) problem. To be specific, given a set
of $n$ training images $\left\{\vec{X}_i\right\}_{i=1}^n$, we
extract $n_i$ object proposals $\{\vec{x}_{ij}, j = 1,\dots,n_i\}$
from each image $\vec{X}_i$ using general object detection
techniques. By decomposing images into object proposals, each
image $\vec{X}_i$ becomes a bag containing several positive
instances, i.e., proposals with the target objects, and negative
instances, i.e., proposals with background or other objects. The
problem of classifying $\vec{X}_i$ is thus transformed from a
multi-label classification problem to a multi-class MIL problem.
The merit of such a transformation is that we do not need to deal
with the complex process of directly recognizing multiple objects
in multiple scales, locations and categories in a single image.
Instead, we only need to identify whether there exist target
objects in the proposals, which has been proven to be the forte of
CNN features~\cite{girshick2014}.

MIL problems assume that every positive bag contains at least one
positive instance. As extensively compared and evaluated
in~\cite{Hosang2014}, state-of-the-art general object detection
methods like BING~\cite{Cheng2014}, selective
search~\cite{Uijlings2013}, MCG~\cite{Arbelaez2014} and
EdgeBoxes~\cite{Zitnick2014} can reach reasonably good recall
rates with several hundreds of proposals. Therefore, if we sample
enough proposals from each image, we can safely assume that these
proposals can cover all objects (or at least all object
categories) in an image, thus fulfilling the assumption of
multi-instance learning.

\begin{figure}
 \centering
 \includegraphics[width=0.35\textwidth]{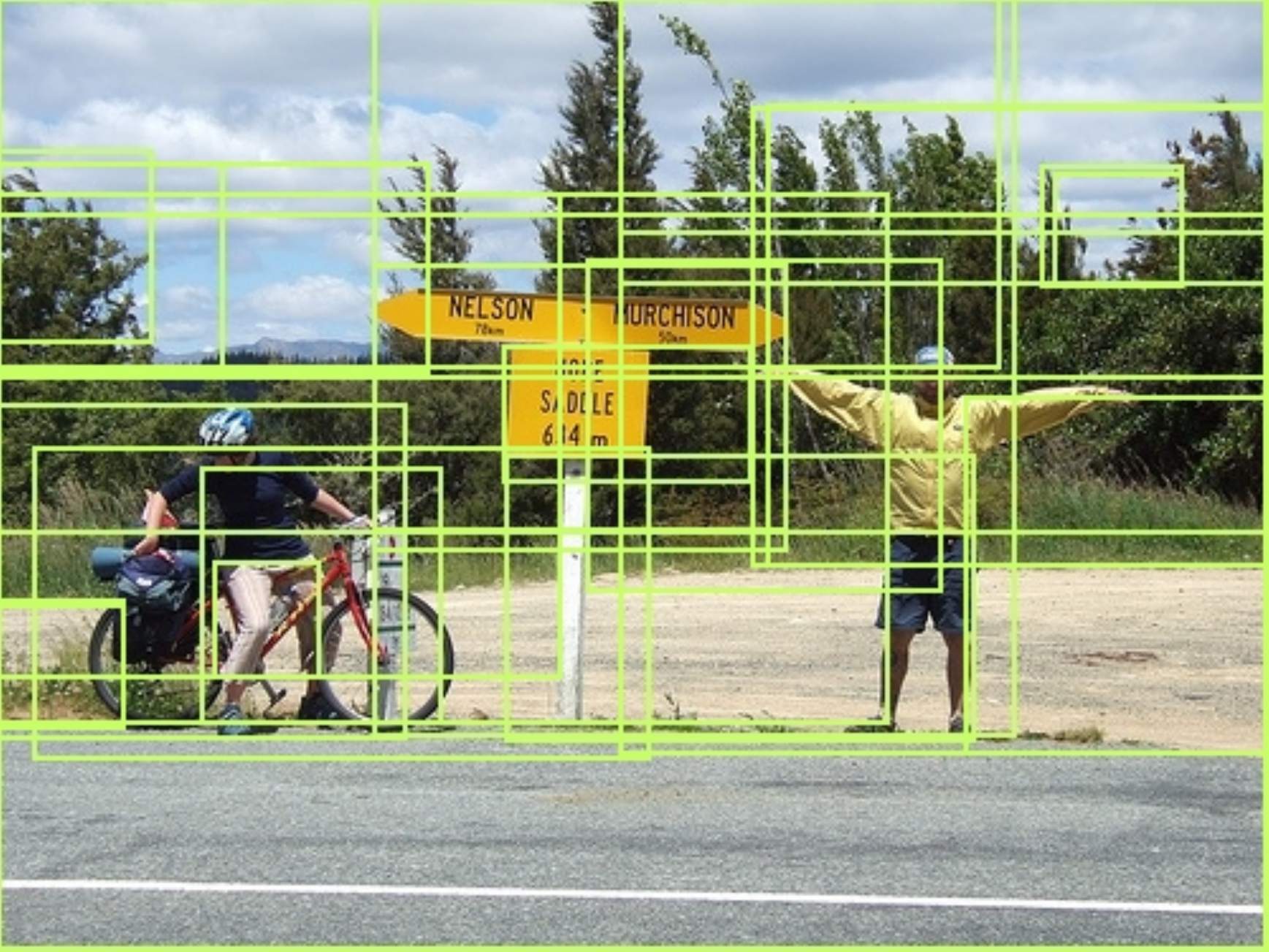}
 \caption{An example of object proposals generated by selective search. We demonstrate $30$ randomly sampled proposals from the full $218$ proposals, which clearly cover two of the main objects: person and bike.}
 \label{selective-search}
\end{figure}

In particular, we employ the unsupervised selective search
method~\cite{Uijlings2013} for object proposal generation.
Selective search has proven to be able to achieve a balance
between effectiveness and efficiency~\cite{Hosang2014}. More
importantly, as it is unsupervised, no extra training data or
ground truth bounding boxes are needed in this stage. Example of
proposals extracted by selective search can be found in
Fig.~\ref{selective-search}.

Traditionally, MIL is formulated as a max-margin classification
problem with latent parameters optimized using alternating
optimization. Typical examples include miSVM~\cite{Andrews2003}
and Latent SVM~\cite{Felzenszwalb2010}. However, although these
methods can achieve satisfactory accuracies, their limitations in
scalability hinder their applicability to current large scale
image classification tasks. For large scale MIL problem,
\cite{Wu2014} shows that Fisher
vector~\cite{Perronnin2010,Sanchez2013} (FV) can be used as an
efficient and effective holistic representation for a bag.
%Moreover, as Fisher vector is originally derived for image classification tasks, it should be more suitable for our task.
Thus, we choose to represent each bag $\vec{X}_i$ as an FV.

Assume we have a $K$-component Gaussian Mixture Model (GMM) with
parameters $\theta = \{\omega_k,\vec{\mu}_k,\Sigma_k,
k=1,\dots,K\}$, where $\omega_k$, $\vec{\mu}_k$ and $\Sigma_k$ are
the mixture weight, mean vector and covariance matrix of the
$k$-th Gaussian, respectively. The covariance matrices $\Sigma_k$
are assumed to be diagonal, where the corresponding standard
deviations of the diagonal entries form a vector $\vec{\sigma}_k$.
We have~\cite{Sanchez2013}:
\begin{align}
%f^{\vec{X}_i}_{\omega_k} =& \frac{1}{\sqrt{\omega_k}}\sum_{j=1}^{n_i}\left(\gamma_j(k)-\omega_k\right) \,, \\
f^{\vec{X}_i}_{\vec{\mu}_k} =& \frac{1}{\sqrt{\omega_k}}\sum_{j=1}^{n_i}\gamma_j(k)\left(\frac{\vec{x}_{ij}-\vec{\mu}_k}{\vec{\sigma}_k}\right) \,, \\
f^{\vec{X}_i}_{\sigma_k} =&
\frac{1}{\sqrt{\omega_k}}\sum_{j=1}^{n_i}\gamma_j(k)\frac{1}{\sqrt{2}}\left[\frac{\left(\vec{x}_{ij}-\vec{\mu}_k\right)^2}{\vec{\sigma}^2_k}
- 1\right] \,,
\end{align}
where $\gamma_j(k)$ is the soft assignment weight, which is also
the probability for $x_{ij}$ to be generated by the $k$-th
Gaussian:
\begin{equation}
\gamma_j(k) =p(k|x_{ij},\theta) .
\end{equation}

We map all the proposals $\{\vec{x}_{ij}, j = 1,\dots,n_i\}$ in an
image $\vec{X}_i$ to an FV by concatenating
$f^{\vec{X}_i}_{\vec{\mu}_k}$ and $f^{\vec{X}_i}_{\vec{\sigma}_k}$
for all $k = 1,\dots,K$, and denoting it as $F^{\vec{X}_i}$.
$F^{\vec{X}_i}$ will be used as the final feature to train the
one-vs-all linear classifiers. Note that for simplicity, we abuse
the notation of $\vec{x}_{ij}$ for both proposal $j$ in image $i$
and its corresponding feature representation. In the next section,
we will describe how to generate the feature representation
$\vec{x}_{ij}$ for each proposal.

\section{From Global Representation to Local Similarity}
\label{lv} Once we obtain object proposals for each image, we can
naturally use CNN features to represent these proposals. Following
general practices in the literature~\cite{Sermanet2013,Razavian2014}, each proposal is fed
into a pre-trained CNN, and the output of the second last fully
connected layer (e.g. layer 7 in AlexNet~\cite{Alex2012}) is used
as the feature representation of that particular proposal. We call
this kind of representation as the feature view $\vec{f}_{ij}$ for
proposal $\vec{x}_{ij}$ from image $\vec{X}_i$. With the proposals
represented by CNN features, one baseline is to encode each image
(bag) at the feature view by Fisher vector as discussed in
Sec.~\ref{mi}. Such baseline is able to get reasonably good
results by utilizing the Fisher vector generated only from the
feature view.

However, since the proposals contain different objects as well as
random background, there exist large variances and imbalanced
distributions. As a consequence, the global representation might
not be accurate enough. Inspired by the idea of local learning,
which solves the data density and intra-class variation problems
by focusing on a subset of the data that are more relevant to a
particular instance~\cite{Bottou1992}, we propose the second level
of locality by adding local spatial configuration information as
the label view (cf. Fig.~\ref{system}) to enhance the
discriminative power of the feature.

To effectively encode the local spatial configuration of a
proposal, we need to solve two key problems: how to form a good
candidate pool for local learning and how to determine which
candidates are relevant to a particular new proposal. For the
former problem, since we have some ground truth object bounding
boxes from the strong labels, we could use them as the candidate
pool assuming all of the ground truth objects are useful. For the
latter one, we follow the common assumption that the most relevant
candidates are the nearest neighbors. In this way, the problem
becomes how to define ``nearest''.  Many
studies~\cite{Bellet2013,Weinberger2009} have shown that the
distance metric is critical to the performances of local learning.

\subsection{CNN as Metric Learning}
\label{largeNN} Metric learning studies the problem of learning a
discriminative distance metric. Conventional Mahalanobis metric
learning methods optimize the parameters of a distance function in
order to best satisfy known similarity or dissimilarity
constraints between training instances~\cite{Bellet2013}. To be
specific, given a set of $n$ labelled training instances
$\{\vec{x}_i, y_i\}_{i=1}^n$, the goal of metric learning is to
learn a square matrix $\mathbf{M}$ such that the distance mapped
between training data, represented as
\begin{equation}
 D_M(\vec{x}_i,\vec{x}_j) = (\vec{x}_i - \vec{x}_j)^T\mathbf{M}(\vec{x}_i - \vec{x}_j) \,,
\end{equation}
satisfies certain constraints. Since $\mathbf{M}$ is symmetric and
positive semi-definite, it can be decomposed as
%\begin{equation}
 $\mathbf{M} = \mathbf{W}^T\mathbf{W}$,
%\end{equation}
and $D_M(\vec{x}_i,\vec{x}_j)$ can be rewritten as:
\begin{equation}
 D(\vec{x}_i,\vec{x}_j) = \left\|\mathbf{W}(\vec{x}_i - \vec{x}_j)\right\|^2 \,.
\end{equation}

We can see that learning a distance metric is equivalent to
learning linear projection $W$ that maps the data from input space
to a transformed space. In this sense, the extraction of CNN
features from the original raw pixel space can also be viewed as a
form of metric learning, while only the process is highly
nonlinear. However, the goal of CNN is usually to minimize the
classification error using loss functions such as the logistic
loss, which may not be suitable for local encoding.

Our desired metric should be discriminative such that all
categories are well separated, as well as compact so that we can
find more accurate nearest neighbors. Specifically, we want the
pairwise distance between instances from the same class to be
smaller than that between instances from different classes. In
order to achieve such a goal,~\cite{Weinberger2009} proposed the
large-margin distance to minimize the following objective
function:
\begin{align}
 \label{objective}
 \sum_{i,j}\eta_{ij}D(\vec{x}_i,\vec{x}_j) +& \notag \\ \alpha \sum_{i,j,l}\eta_{ij}(1-y_{il}) \bigl [1 + D(\vec{x}_i,\vec{x}_j) - D(\vec{x}_i,\vec{x}_l) \bigr ]_{+} & \,.
\end{align}
Here $\eta$ encodes target nearest neighbor information, where
$\eta_{ij} = 1$ if $\vec{x}_j$ is one of the $\hat{k}$ positive
nearest neighbors of $\vec{x}_i$; otherwise $\eta_{ij} = 0$. $y$
is the label information where $y_{il} = 1$ if  $\vec{x}_i$ and
$\vec{x}_l$ are in the same class; otherwise $y_{il} = 0$.
$[\cdot]_+ = \max(\cdot,0)$ is the hinge loss function. $\alpha$
is the trade-off parameter. The first term in Eq.~\ref{objective}
penalizes large distances between instances and target neighbors,
and the second term penalizes small distances between each
instance and all other instances that do not share the same label.
By employing such an objective function, we can ensure that the
$\hat{k}$-nearest neighbors of an instance belong to the same
class, while instances from different classes are separated by a
large margin.

In order to learn a discriminative metric, we propose to learn a
large-margin nearest neighbor (LMNN) CNN. Specifically, we replace
the logistic loss with the large margin nearest neighbor loss and
train a network with low-dimension output utilizing the strong
labels. Details of training and fine-tuning the LMNN CNN can be
found in Section~\ref{details}. The output of the proposed LMNN
network is a low-dimensional feature that shares the good semantic
abstraction power of conventional CNN feature and the good
neighborhood property of large-margin metric learning. We then
build the candidate pool with the LMNN CNN features extracted from
ground truth objects.

\subsection{Encoding Local Spatial Label Distribution as Label View} \label{encode}
To effectively incorporate local label information around a local
region, we encode its neighborhood as the label view.
Specifically, we extract features from each proposal
$\vec{x}_{ij}$ using the LMNN CNN, then find $k$ nearest neighbors
of $\vec{x}_{ij}$ in the candidate feature pool as $\vec{nn}_{ij}
= \{nn^1_{ij}, \dots,nn^k_{ij}\}$ and record their labels
$\vec{l}_{ij} = \left[l^1_{ij}\text{ }\dots\text{ }
l^k_{ij}\right]$. The label information (e.g. $l^k_{ij}$) of a
neighbor (e.g. $nn^k_{ij}$) is encoded as a $C$-dimensional binary
vector, which corresponds to $C$ categories. The $d$-th dimension
$l^k_{ij}(d) = 1$, $(d =1,\dots, C)$ if the object is annotated as
class $d$; otherwise $l^k_{ij}(d) = 0$. Therefore, $\vec{l}_{ij}$
is a $1 \times kC$ vector and it will be used as the feature for
the label view.

The merit of such indirectly utilizing ground truth bounding boxes
as the label view is the good generalization ability. As the label
view is a form of local structure representation, even for unseen
categories, i.e. no same-category strong labels, the encoding
process can naturally exploit existing semantically or visually
close categories to build local support. For example, suppose we
do not have the bounding box annotations for ``cat'' and
``train'', a proposal containing ``cat'' might have nearest
neighbors of ``dog'', ``tiger'' or other related animals, and a
proposal containing ``train'' might have nearest neighbors of
``car'', ``truck'' or other vehicles. Although lacking the exact
annotations of certain category, the label view is still able to
encode the local structure with semantically or visually similar
objects. In this way, our framework can make use of existing
strong supervision information to boost the overall performance.
The experimental results shown in Section~\ref{results} validate
this argument.

We directly concatenate the feature view and the label view to
form the final representation of each proposal $\vec{x}_{ij}$ as
$\left[\vec{f}_{ij}\text{ }\lambda \vec{l}_{ij}\right]$, where
$\lambda$ is the trade-off parameter between the feature view and
the label view.

\subsection{Network Configurations and Implementation Details} \label{details}

Our framework consists of two networks, a large-margin nearest
neighbor (LMNN) CNN and a standard CNN. Both networks'
architectures are similar to~\cite{Chatfield2014} with $5$
convolutional layers and $3$ fully-connected layers, and the
dimension of the layer-7 output is set to 2048. The main
differences of these two networks lie in the loss function and the
fine-tuning process. For LMNN CNN, its layer-8 output is a
128-dimensional feature, based on which we measure the pair-wise
distance for kNN, while the output of the standard CNN is a
$C$-dimensional score vector, corresponding to the $C$ categories.

\textbf{Data pre-processing and pre-training.} We use the ILSVRC
$2012$ dataset to pre-train both networks. Given an image, we
resize the short side to $224$ with bilinear interpolation and
perform a center crop to generate the standard $224 \times 224$
input. Each of these inputs is then pre-processed by subtracting
the mean of ILSVRC images.

\textbf{Fine-tuning.} To better adopt the pre-trained network for
specific applications, we also fine-tune these networks using task
relevant data. Unlike~\cite{Chatfield2014}, currently our
implementation does not involve any data augmentation in the
fine-tuning stage.

For the standard CNN used for feature view, we only fine-tune the
network with weak labels on the whole image. As our task is
multi-label recognition, following~\cite{Wei2014}, we use square
loss instead of the logistic loss. To be specific, suppose we have
a label vector $\vec{y}_i = \left[y_{i1}, y_{i2},\dots,
y_{iC}\right]$ for the $i$-th image. $y_{ij} = 1$ $(j =1,\dots,
C)$ if the image is annotated with class $j$; otherwise $y_{ij} =
0$. The ground truth probability vector of the $i$-th image is
defined as $\vec{p}_i = \vec{y}_i/\left\|\vec{y}_i\right\|_1$ and
the predicted probability vector is $\hat{\vec{p}}_i =
\left[\hat{p}_{i1}, \hat{p}_{i2},\dots, \hat{p}_{iC}\right]$.
Then, the cost function to be minimized is defined as
\begin{equation}
 \frac{1}{n}\sum_{i=1}^{n}\sum_{j=1}^{C}(p_{ij} - \hat{p}_{ij})^2 \,.
\end{equation}
During the fine-tuning, the parameters of the first seven layers
of the network are initialized with the pre-trained parameters.
The parameters of the last fully connected layer is initialized
with a Gaussian distribution. We tune the network for $10$ epochs
in total.

For the large-margin NN (LMNN) CNN used for label view, we execute
a three-step fine-tuning. The first step is the image level
fine-tuning similar to the process we have described above. The
second step is ground truth objects fine-tuning, where we
fine-tune the network using ground truth objects with the logistic
loss. The final step is the large-margin nearest neighbor
fine-tuning, where we fine-tune the network with the loss function
of Eq.~\ref{objective} described in Section~\ref{largeNN}. To
accelerate the process, in this final step, we fix all parameters
of the first seven layers and only fine-tune the parameters of the
last fully connected layer.

\section{Experimental Results} \label{exp}

In this section, we present the experimental results of the
proposed multi-view multi-instance framework on multi-label object
recognition tasks.

\subsection{Datasets and Baselines}

We evaluate our method on the PASCAL Visual Object Classes
Challenge (VOC) datasets~\cite{VOC}, which are widely used as
benchmark datasets for the multi-label object recognition task. In
particular, we use the VOC 2007 and VOC 2012 datasets. The details
of these datasets can be found in Table~\ref{datasets}. These two
datasets have a pre-defined split of \textsc{train}, \textsc{val}
and \textsc{test} sets. We use \textsc{train} and \textsc{val} for
training and \textsc{test} for testing. The evaluation method is
average precision (AP) and mean average precision (mAP).

\begin{table}
 \centering
 \caption{Dataset information.} \label{datasets} \vspace{0.1in}
 \begin{tabular}{c | r r r }
  \hline
  Dataset &\#TrainVal &\#Test &\#Classes \\ \hline
  \textsc{VOC 2007} &5011 &4952 &20 \\
  \textsc{VOC 2012} &11540 &10991 &20 \\ \hline
 \end{tabular}
\end{table}

We compare the proposed framework with the following
state-of-the-art approaches: \squishlist
 \item CNN-SVM~\cite{Razavian2014}. This method employed OverFeat~\cite{Sermanet2013}, which is pre-trained on ImageNet, to get CNN activations as the off-the-shelf features. Specifically, CNN-SVM employs the $4096$-d feature extracted from the $22$-nd layer of OverFeat and uses these features to train a linear SVM for the classification task.

 \item PRE~\cite{OquabCVPR2014}.  \cite{OquabCVPR2014} proposed to transfer image representations learned with CNN on ImageNet to other visual recognition tasks with limited training data. The network has exactly the same architecture as that in~\cite{Alex2012}. The network is first pre-trained on ImageNet. The parameters of the first seven layers of CNN are then fixed and  the last fully-connected layer is replaced by two adaptation layers. Finally, the adaptation layers are trained with images from the target dataset.

 \item HCP~\cite{Wei2014}. HCP proposed to solve the multi-label object recognition task by extracting object proposals from the images. Specifically, HCP has three main steps. The first step is to pre-train a CNN on ImageNet data. The second step is image-level fine-tuning that uses image labels and square loss to fine-tune the pre-trained CNN. The final step is to employ BING~\cite{Cheng2014} to extract object proposals and fine-tune the network with these proposals. The image-level scores are obtained by max-pooling from the scores of the proposals.

 \item~\cite{Oquab2014}. \cite{Oquab2014} also handled the problem
in a weakly supervised manner. Particularly, multiple windows are
extracted from different scales of the images in the dense
sampling fashion. The scores of these windows are combined with
max-pooling from the same scale then sum-pooling across different
scales.

 \item VeryDeep~\cite{Simonyan2014}. \cite{Simonyan2014} densely extracts multiple CNN features across multi-scales of the image with very-deep networks (16-layer and 19-layer). The features from the same scale are concatenated by sum-pooling and features from different scale are aggregated by stacking or sum-pooling.  \cite{Simonyan2014} also augments the test set by horizontal flipping of the images.

 \item Hand-crafted Features. \cite{AMM2014} presented an
Ambiguity guided Mixture Model (AMM) to integrate external context
features and object features, and then used the contextualized SVM
to iteratively boost the performance of object classification and
detection tasks.~\cite{Dong2013} proposed an Ambiguity Guided
Subcategory (AGS) mining approach to improve both detection and
classification performance. \squishend

\subsection{Our Setup and Parameters}
It is difficult to make a completely fair comparison among
different CNN based methods as the CNN configurations, the data
augmentation and the pre-training could substantially influence
the results. All CNN based methods can benefit from extra training
data and more powerful networks as shown
in~\cite{OquabCVPR2014,Wei2014,Oquab2014,Chatfield2014,Simonyan2014}.
To fairly evaluate our proposed framework, we develop our system
based on the common $8$-layer CNN pretrained on ILSVRC 2012
dataset with 1000 categories. The details of the fine tuning
process has been elaborated in Section~\ref{details}. Once the
LMNN CNN and the standard CNN (see Fig.~\ref{system}) are trained,
the system is applied to map each training image into a final FV
feature. Finally, TopPush~\cite{Li2014} is chose to learn linear
one-vs-all classifiers for each category, which produces the
scores for each binary sub-problem of the multi-label datasets.
The scores are then evaluated with standard VOC evaluation
package. All the experiments are run on a computer with Intel
i7-3930K CPU, 32G main memory and an nVIDIA Tesla K40 card.

For the proposal extraction, we employ selective
search~\cite{Uijlings2013}, which typically generates around
$1500$ proposals on average from every image in the PASCAL VOC
2007 dataset using the parameters suggested
in~\cite{Uijlings2013}. Considering the computational time and the
hardware limitation, we random sample around $400$ proposals per image for training and testing. 

For the parameters of Fisher vector, we
follow~\cite{Perronnin2010} to first employ PCA to reduce the
dimension of the original features to preserve around 90\% energy.
For VOC 2007 and 2012 datasets, after PCA, the standard CNN
features is reduced to around $450$-d. After PCA, we generate $128$
GMM codewords and encode each image with IFV similar to~\cite{Perronnin2010}.

For fine-tuning the LMNN CNN, we set the trade-off parameter
$\alpha=1$ (see~\eqref{objective}) and the nearest neighbor number
$\hat{k}=10$ (for training). For combining the feature view and
the label view features, we select the trade-off parameter
$\lambda$ (specified at the end of Section~\ref{encode}) from
$\{1,0.5,0.25\}$ by cross-validation. For the nearest neighbor
number $k$ used in testing, generally bigger $k$ leads to better
accuracy, but we observe there is no performance gain for $k >
50$. Thus, we set $k=50$ for testing. For faster NN search, we
employ FLANN~\cite{flann} with ``autotuned'' parameters.

\subsection{Image Classification Results} \label{results}

\begin{table*}
 \centering \caption{Comparisons of the classification results (in
 \%) of state-of-the-art approaches on VOC 2007
 (\textsc{trainval}/\textsc{test}). The upper part shows the
 results of the hand-crafted feature based methods and the CNN based methods
trained with $8$-layer CNN and ILSVRC $2012$ dataset. The lower
part shows the results of the methods trained with very-deep CNN
or with additional training data.} \label{voc2007-s} \scriptsize
 \begin{tabular}{ @{\,}c| *{20}{@{\,}c@{\,}} | @{\,}c@{\,}}
 \hline
                                &\textsc{plane} &\textsc{bike} &\textsc{bird} &\textsc{boat}   &\textsc{bottle} &\textsc{bus}  &\textsc{car}   &\textsc{cat}  &\textsc{chair} &\textsc{cow}
                                &\textsc{table} &\textsc{dog}  &\textsc{horse} &\textsc{motor} &\textsc{person} &\textsc{plant} &\textsc{sheep} &\textsc{sofa}  &\textsc{train} &\textsc{tv}    &\textsc{mAP}\\ \hline
  AGS~\cite{Dong2013}           &82.2  &83.0 &58.4 &76.1   &56.4   &77.5 &88.8  &69.1 &62.2 &61.8  &64.2 &51.3  &85.4  &80.2  &91.1   &48.1  &61.7  &67.7  &86.3  &70.9  &71.1\\
  AMM~\cite{AMM2014}            &84.5  &81.5 &65.0 &71.4   &52.2   &76.2 &87.2  &68.5 &63.8 &55.8  &65.8 &55.6  &84.8  &77.0  &91.1   &55.2  &60.0  &69.7  &83.6  &77.0  &71.3\\
  CNN-SVM~\cite{Razavian2014}   &88.5  &81.0 &83.5 &82.0   &42.0   &72.5 &85.3  &81.6 &59.9 &58.5  &66.3 &77.8  &81.8  &78.8  &90.2   &54.8  &71.1  &62.6  &87.4  &71.8  &73.9\\
  PRE-1000C~\cite{OquabCVPR2014}&88.5  &81.5 &87.9 &82.0   &47.5   &75.5 &90.1  &87.2 &61.6 &75.7  &67.3 &85.5  &83.5  &80.0  &95.6   &60.8  &76.8  &58.0  &90.4  &77.9  &77.7\\
  HCP-1000C~\cite{Wei2014}      &95.1  &90.1 &92.8 &89.9   &51.5   &80.0 &91.7  &91.6 &57.7 &77.8  &70.9 &89.3  &89.3  &85.2  &93.0   &64.0  &\bf{85.7}  &62.7  &94.4  &78.3  &81.5\\ \hline
  FeV                            &93.3  &92.8 &91.8 &86.5   &57.5   &84.3 &93.7  &90.6 &64.7 &78.9  &74.1 &90.3  &91.1  &90.7  &95.7   &67.4  &82.4  &70.8  &94.3  &83.3  &83.7\\
  FeV+LV-10                      &\bf{96.6} &93.6 &\bf{94.0} &\bf{89.5}   &59.6   &87.4 &\bf{94.8}  &91.0 &67.3 &81.4  &76.3 &91.0  &\bf{93.5}  &91.4  &96.1   &64.2  &83.4  &68.5  &95.8  &84.0  &85.0\\
  FeV+LV-20                      &95.7  &\bf{94.6} &93.9 &87.8   &\bf{62.7}   &\bf{87.9} &\bf{94.8}  &\bf{92.2} &\bf{67.5} &\bf{82.4}  &\bf{77.8} &\bf{92.0}  &93.2  &\bf{92.2}  &\bf{97.1}   &\bf{72.5}  &85.3  &\bf{73.4}  &\bf{96.2}  &\bf{85.5}
                                &\bf{86.2}\\
  \hline
  \hline
  HCP-2000C~\cite{Wei2014}  &96.0  &92.1 &93.7 &93.4   &58.7   &84.0 &93.4  &92.0 &62.8 &89.1  &76.3 &91.4  &95.0  &87.8  &93.1   &69.9  &90.3  &68.0  &96.8  &80.6  &85.2\\
  VeryDeep~\cite{Simonyan2014}  &\bf{98.9}  &95.0 &96.8 &95.4   &69.7   &90.4 &93.5  &96.0 &74.2 &86.6  &87.8 &96.0  &96.3  &93.1  &97.2   &70.0  &92.1  &80.3  &98.1  &87.0  &89.7\\ \hline
  FeV+LV-20-VD           &97.9  &\bf{97.0} &96.6 &94.6   &73.6   &93.9 &96.5  &95.5 &73.7 &90.3  &82.8 &95.4  &\bf{97.7}  &\bf{95.9}  &\bf{98.6}   &77.6  &88.7  &78.0  &98.3  &89.0  &90.6\\
  Fusion            &98.2  &96.9 &\bf{97.1} &\bf{95.8}   &\bf{74.3}   &\bf{94.2} &\bf{96.7}  &\bf{96.7} &\bf{76.7} &\bf{90.5}  &\bf{88.0} &\bf{96.9}  &\bf{97.7}  &\bf{95.9}  &\bf{98.6}   &\bf{78.5}  &\bf{93.6}  &\bf{82.4}
  &\bf{98.4}  &\bf{90.4}  &\bf{92.0}\\
  \hline
 \end{tabular}
\end{table*}

\begin{table*}
 \centering \caption{Comparisons of the classification results (in
 \%) of state-of-the-art approaches on VOC 2012
 (\textsc{trainval}/\textsc{test}). The upper part shows the
 results of the hand-crafted feature based methods and the CNN based methods
trained with $8$-layer CNN and ILSVRC $2012$ dataset. The lower
part shows the results of the methods trained with very-deep CNN
or with additional training data.} \label{voc2012-s} \scriptsize
 \begin{tabular}{ @{\,}c| *{20}{@{\,}c@{\,}} | @{\,}c@{\,}}
 \hline
                                &\textsc{plane} &\textsc{bike} &\textsc{bird} &\textsc{boat}   &\textsc{bottle} &\textsc{bus}  &\textsc{car}   &\textsc{cat}  &\textsc{chair} &\textsc{cow}
                                &\textsc{table} &\textsc{dog}  &\textsc{horse} &\textsc{motor} &\textsc{person} &\textsc{plant} &\textsc{sheep} &\textsc{sofa}  &\textsc{train} &\textsc{tv}    &\textsc{mAP}\\ \hline
  NUS-PSL~\cite{Wei2014}    &97.3  &84.2 &80.8 &85.3   &60.8   &89.9 &\bf{86.8}  &89.3 &\bf{75.4} &77.8  &75.1 &83.0  &87.5  &90.1  &95.0   &57.8  &79.2  &\bf{73.4}  &94.5  &80.7  &82.2\\
  PRE-1000C~\cite{OquabCVPR2014}&93.5  &78.4 &87.7 &80.9   &57.3   &85.0 &81.6  &89.4 &66.9 &73.8  &62.0 &89.5  &83.2  &87.6  &95.8   &61.4  &79.0  &54.3  &88.0  &78.3  &78.7\\
  HCP-1000C~\cite{Wei2014}      &\bf{97.7}  &83.0 &\bf{93.2} &87.2   &59.6   &88.2 &81.9  &94.7 &66.9 &81.6  &68.0 &\bf{93.0}  &88.2  &87.7  &92.7   &59.0  &85.1  &55.4  &93.0  &77.2  &81.7\\ \hline
  FeV                            &96.8  &87.8 &88.7 &87.2   &63.8   &\bf{92.3} &86.2  &92.3 &72.4 &82.0  &76.0 &91.9  &90.3  &90.3  &95.2   &61.2  &82.6  &65.6  &92.8  &84.4  &84.0\\
  FeV-LV-10                      &97.3  &\bf{89.1} &\bf{91.5} &\bf{88.5}   &\bf{66.7}   &92.2 &\bf{87.2}  &94.0 &74.0 &82.7  &77.8 &91.6  &91.1  &92.7  &95.7   &66.5  &85.4  &\bf{69.4}  &\bf{95.6}  &85.4  &85.7\\
  FeV-LV-20                      &\bf{97.4}  &88.9 &91.2 &87.4   &64.2   &92.2 &86.4  &\bf{95.0} &\bf{75.1} &\bf{84.6}  &\bf{78.7} &\bf{93.1}  &\bf{91.9}  &\bf{93.1}  &\bf{96.6}   &\bf{67.3}  &\bf{86.2}  &\bf{69.4}  &95.3  &\bf{85.8}  &\bf{86.0}\\
  \hline
  \hline
  PRE-1512C~\cite{OquabCVPR2014}&94.6  &82.9 &88.2 &84.1   &60.3   &89.0 &84.4  &90.7 &72.1 &86.8  &69.0 &92.1  &93.4  &88.6  &96.1   &64.3  &86.6  &62.3  &91.1  &79.8  &82.8\\
  HCP-2000C~\cite{Wei2014}      &97.5  &84.3 &93.0 &89.4   &62.5   &90.2 &84.6  &94.8 &69.7 &90.2  &74.1 &93.4  &93.7  &88.8  &93.3   &59.7  &90.3  &61.8  &94.4  &78.0  &84.2\\
  \cite{Oquab2014}      &96.7  &88.8 &92.0 &87.4   &64.7   &91.1 &87.4  &94.4 &74.9 &89.2  &76.3 &93.7  &95.2  &91.1  &97.6   &66.2  &91.2  &70.0  &94.5  &83.7  &86.3\\
  VeryDeep~\cite{Simonyan2014}  &99.0  &89.1 &\bf{96.0} &\bf{94.1}   &74.1   &92.2 &85.3  &\bf{97.9} &79.9 &92.0  &\bf{83.7} &\bf{97.5}  &96.5  &94.7  &97.1   &63.7  &\bf{93.6}  &75.2  &97.4  &87.8  &89.3\\
  NUS-HCP-AGS~\cite{Wei2014}    &\bf{99.0}  &91.8 &94.8 &92.4   &72.6   &\bf{95.0} &\bf{91.8}  &97.4 &\bf{85.2} &\bf{92.9}  &83.1 &96.0  &96.6  &\bf{96.1}  &94.9   &68.4  &92.0  &\bf{79.6}  &97.3  &88.5  &90.3\\ \hline
  FeV+LV-20-VD           &98.4  &92.8 &93.4 &90.7   &74.9   &93.2 &90.2  &96.1 &78.2 &89.8  &80.6 &95.7  &96.1  &95.3  &97.5   &73.1  &91.2  &75.4  &97.0  &88.2  &89.4\\
  Fusion            &98.9  &\bf{93.1} &\bf{96.0} &\bf{94.1}   &\bf{76.4}   &93.5 &90.8  &\bf{97.9} &80.2 &92.1  &82.4 &97.2  &\bf{96.8}  &95.7  &\bf{98.1}   &\bf{73.9}  &\bf{93.6}  &76.8  &\bf{97.5}  &\bf{89.0}  &\bf{90.7}\\
  \hline
 \end{tabular}
\end{table*}

\textbf{Image Classification on VOC 2007}: Table~\ref{voc2007-s}
reports our experimental results compared with state-of-the-art
methods on VOC 2007. In the upper part of the table we compare
with the hand-crafted feature based methods and the CNN based
methods pre-trained on ILSVRC 2012 using $8$-layer network. To
demonstrate the effectiveness of individual components, we
consider three variations of our proposed framework: `FeV',
`FeV+LV-10' and `FeV+LV-20', where `FeV' uses only the feature
view (i.e. without the label view features), `FeV+LV-10' uses both
the feature view and the label view with 10 categories of
ground-truth bounding boxes of the training set, and `FeV+LV-20'
is the one with 20 categories of ground-truth bounding boxes.

From the upper part of Table~\ref{voc2007-s}, we can see that
using just feature view (`FeV'), we already outperform the
state-of-the-art proposal-based method (`HCP-1000C') by $2.2\%$,
which suggests that Fisher vector as a holistic representation for
bags is superior than max-pooling. With all 20 categories of
ground-truth bounding boxes of the training set, our multi-view
framework (`FeV+LV-20') achieves a further $2.5\%$ performance
gain. This significant performance gain validates the
effectiveness of the label view. Our framework shows good
performance especially for difficult categories such as
\textsc{bottle}, \textsc{cow}, \textsc{table}, \textsc{motor} and
\textsc{plant}.

If we just use the ground-truth bounding boxes from the first $10$
categories (\textsc{plane} to \textsc{cow}), our framework
(`FeV+LV-10') still outperforms single feature view (`FeV') by a
margin of $1.3\%$. As expected, using the bounding boxes of the
categories from \textsc{plane} to \textsc{cow} can boost the
performance of these categories as shown in the table. However, it
is interesting to see that the label view also improves the
accuracies of unseen categories such as \textsc{horse},
\textsc{person} and \textsc{tv}. This is mainly because the
proposed label view encoding is a form of local similarity
representation, which can generalize quite well to unseen
categories.

In the lower part of Table~\ref{voc2007-s}, we list the results of
`HCP-2000C'~\cite{Wei2014}, which uses additional $1000$
categories from ImageNet that are semantically close to VOC 2007
categories for CNN pre-training, and
`VeryDeep'~\cite{Simonyan2014}, which densely extracts multiple
CNN features from $5$ scales and combines two very-deep CNN models
($16$-layer an $19$-layer). Our framework (`FeV+LV-20') can still
outperform `HCP-2000C', but is inferior to `VeryDeep' since our
framework is based on the common 8-layer CNN.

To demonstrate the potential of our framework, we replace the
8-layer CNNs in our framework by the 16-layer CNN model
in~\cite{Simonyan2014}, which is denoted as `FeV+LV-20-VD'.
Unlike~\cite{Simonyan2014}, we do not use any data augmentation or
multi-scale dense sampling in the feature extraction stage. Our
`FeV+LV-20-VD' outperforms `VeryDeep' by nearly $1\%$. By further
averaging the scores of `VeryDeep'~\cite{Simonyan2014} and
`FeV+LV-20-VD' (denoted as `Fusion'), we achieve state-of-the-art
mAP of $92.0\%$. This suggests that our proposal-based framework
and the multi-scale CNN extracted from the whole image are
complement to each other.

\textbf{Image Classification on VOC 2012}: Table~\ref{voc2012-s}
reports our experimental results compared with those of the
state-of-the-art methods on VOC 2012. Similar to
Table~\ref{voc2007-s}, we compare with the hand-crafted feature
based methods and the CNN based methods pre-trained on ILSVRC 2012
using $8$-layer CNN model in the upper part and the methods
trained with additional data or very-deep CNN models in the lower
part.

The results are consistent with those on VOC 2007. Our framework
that uses only the feature view (`FeV') already outperforms the
state-of-the-art hand-crafted feature method (`NUS-PSL') by
$1.8\%$ and the state-of-the-art proposal-based CNN method
(HCP-1000C) by $2.3\%$. With the aid of the label view, our
`FeV+LV-20' obtains an additional $2\%$ performance again, even
outperforming the two proposal-based methods pre-trained on
additional $512$ or $1000$ categories of image data (`PRE-1512C'
and `HCP-2000C') and comparable to~\cite{Oquab2014}. By employing
just $10$ categories of bounding boxes, the mAP performance of our
`FeV+LV-10' does not degrade much.

When employed with the very-deep 16-layer CNN
model~\cite{Simonyan2014}, our framework (`FeV+LV-20-VD') achieves
similar performance as `VeryDeep'. When we averagely fuse the
scores of~\cite{Simonyan2014} and our proposal-based
representation, our method (`Fusion') achieves state-of-the-art
result, outperforming~\cite{Wei2014}.
%Note that we directly apply $0.5$ as the trade-off parameter $\lambda$ in the fusion stage, and we believe tuning this parameter can further improve the performance of the framework.

\section{Conclusion}

In this paper, we have proposed a multi-view multi-instance
framework for solving the multi-label classification problem.
Compared with existing works, our framework makes use of the strong labels to provide another view of local information (label view) and combines it with the typical feature view information to boost the discriminative power of feature
extraction for multi-label images. 
The experimental results validates the discriminative power and the
generalization ability of the proposed framework.

For future directions, there are several possibilities to explore.
First of all, we can improve the scalability and possibly also
the performance of the framework by establishing a
proposal selection criteria to filter out noisy proposals.
Secondly, we may build a suitable candidate pool directly from the
extracted proposals to eliminate the need for strong labels.

\noindent\textbf{Acknowledgments} \small This research is supported by Singapore MoE AcRF Tier-1 Grant
RG138/14 and also partially supported by the Rapid-Rich Object
Search (ROSE) Lab at the Nanyang Technological University,
Singapore. The Tesla K40 used for this research was donated by the NVIDIA Corporation.

{\small
\bibliographystyle{ieee}
\bibliography{egbib}
}

\end{document}